\documentclass[letterpaper]{article} 
\usepackage{aaai2026}  
\makeatletter
\def\@copyrightspace{} 
\def\copyright@on{}    
\makeatother

\usepackage{times}  
\usepackage{amssymb}  
\usepackage{helvet}  
\usepackage{courier}  
\usepackage{amsmath}  
\usepackage[hyphens]{url}  
\usepackage{graphicx} 
\usepackage{natbib}  
\usepackage{caption} 
\usepackage{algorithm}
\usepackage{algorithmic}

\title{Accurate Predictions in Education with Discrete Variational Inference}

\title{Accurate Predictions in Education with Discrete Variational Inference}
\author {
    Tom Quilter\textsuperscript{\rm 1}, 
    Anastasija Ilick\textsuperscript{\rm 2}, 
    Karen Poon\textsuperscript{\rm 3}, 
    Richard Turner\textsuperscript{\rm 4}
}
\affiliations {
    \textsuperscript{\rm 1}University of Manchester \\
    \textsuperscript{\rm 2}Google DeepMind \\
    \textsuperscript{\rm 3}MathWorks \\
    \textsuperscript{\rm 4}University of Cambridge \\
    \vspace{0.5em}
    \textit{All correspondence, \texttt{thomas.quilter@manchester.ac.uk}}
}

\begin{document}
\maketitle

\begin{abstract}
One of the largest drivers of social inequality is unequal access to personal tutoring, with wealthier individuals able to afford it, while the majority cannot. Affordable, effective AI tutors offer a scalable solution. We focus on adaptive learning, predicting whether a student will answer a question correctly, a key component of any effective tutoring system. Yet many platforms struggle to achieve high prediction accuracy, especially in data-sparse settings. To address this, we release the largest open dataset of professionally marked formal mathematics exam responses to date. We introduce a probabilistic modelling framework rooted in Item Response Theory (IRT) that achieves over 80 percent accuracy, setting a new benchmark for mathematics prediction accuracy of formal exam papers. Extending this, our collaborative filtering models incorporate topic-level skill profiles, but reveal a surprising and educationally significant finding, a single latent ability parameter alone is needed to achieve the maximum predictive accuracy. Our main contribution though is deriving and implementing a novel discrete variational inference framework, achieving our highest prediction accuracy in low-data settings and outperforming all classical IRT and matrix factorisation baselines.
\end{abstract}

\section{Introduction}\label{sec:intro}

Everyone remembers an amazing teacher from their childhood. Such individuals possess a passion for their subject and the ability to convey complex ideas. However, one of the most important and sometimes overlooked skills that incredible teachers have is to be able to present the student's work at exactly the right level. This principle is echoed in Duolingo’s design philosophy, where the platform aims for learners to answer around 80 percent of questions correctly. This level of difficulty is challenging enough to sustain engagement, but not so difficult as to become discouraging, occupying the so-called Goldilocks zone of learning (\cite{bjork1994memory}, \cite{csikszentmihalyi1990flow}). 

AI personal tutors hold the promise of helping to redress educational inequality, an inequality driven in large measure by unequal access to high quality personal tutoring. Although wealthier families can afford such individualised support, the majority of students cannot, exacerbating long-standing social divides. For AI tutoring platforms to succeed, their effectiveness will in part depend on whether they can deliver questions at just the right level to keep students engaged and stretched, just as the very best human tutors do.

Our aim in this paper is simply to predict whether a student will get an unseen question correct, given their previous question answers, other students' responses, and meta-data. We hereby release the largest open benchmark of marked mathematics exam paper responses to date, and create models to predict a student’s question success. Moreover, educational platforms regularly have the problem of limited data, limited by the number of students for new platforms and for all platforms limited by the number of questions brand new students have answered. We produce novel methods to predict accurately in these limited data scenarios.  

A natural place to start is a model incorporating the ability of the student and the difficulty of a question, a two-parameter model. We achieve remarkable accuracy with this simple ability-difficulty model. So much so that when we introduce a more complex interaction model allowing for students' individual strengths and weaknesses in certain topics of mathematics, such as algebra, data and geometry, we are unable to beat it. This suggests a potentially important result for education: that a single latent ability parameter may be the main factor driving a student’s success on exam questions, and perhaps the only one needed.  In our search for factors that improve prediction accuracy, we find some benefit in modelling shared topic-level strengths and weaknesses across students in the same high school class, often taught by a single teacher. We also uncover interesting interpretability results, showing that the model implicitly discovers meaningful question types such as algebra or geometry as a byproduct of its optimisation for predictive accuracy.

A major challenge for new educational platforms is the cold-start problem, having too few students on the system to make accurate predictions. It is in this low data regime that we make our most significant breakthrough. We derive a novel discrete variational inference framework, placing Gaussian priors over latent ability variables, which outperforms all our other models when student data is limited. For completeness, we also address the challenge of predicting performance for new students who have only attempted a small number of questions. We apply pool-based active learning to achieve significantly higher predictive accuracy than random selection and we also identify which questions are most informative to present, guiding personalised assessment.

The main contributions of our work include: 

\begin{itemize}
  \item \textbf{When data are scarce our novel Variational Inference for discrete data yields improved accuracy } Placing uncertainty over all latent parameters delivers our most important results.
    \item \textbf{Open-sourcing a New Large-Scale High Quality Educational Dataset} We release the largest open dataset of professionally marked formal mathematics exam responses to date, providing a valuable benchmark.  
  \item \textbf{Ability is all you need:} Remarkably we discover that a single parameter of student ability cannot be beaten by more complex models incorporating individual topic strengths and weaknesses.

\end{itemize}

\section{Related Work}

\paragraph{Item Response Theory (IRT) in Education.}
IRT offers a probabilistic framework for modeling the probability that a learner answers an item correctly as a function of latent traits.  
The simplest variant—the Rasch 1PL model—uses a single student‐ability and item‐difficulty parameter with a logistic link \cite{Rasch1960}.  
Classic extensions add item discrimination and guessing parameters~\citep{Birnbaum1968}.  
Foundational psychometric texts firmly established IRT for large‑scale assessment \citep{Hambleton1991}.  
Early educational technology work showed its practical value: \citet{Chen2006} used IRT to adapt e‑learning content difficulty and demonstrated improved engagement.  
Modern computerised adaptive testing likewise selects questions in real time based on IRT estimates to maximise information gain.

\paragraph{Collaborative Filtering and Matrix Factorisation.}
Viewing students as “users’’ and questions as “items’’ yields a natural analogy to recommender systems.  
Matrix‑factorization (MF) methods learn low‑rank embeddings for students and items and, with a logistic link, are algebraically similar to Rasch.   
MF has repeatedly outperformed feature‑engineered approaches on benchmarks such as the KDD Cup 2010; e.g.\ \citet{ThaiNghe2012} reported lower prediction error than per‑skill models.  
MF also generalises to grade prediction across courses \citep{Polyzou2016}.  
Its flexibility allows side features or temporal components via factorization machines or tensor factorization.  
Notably, \citet{Vie2019} introduced \emph{Knowledge Tracing Machines}, unifying IRT, logistic regression and performance‑factor models in a factorization‑machine framework that achieved state‑of‑the‑art accuracy while remaining interpretable.

\paragraph{Knowledge Tracing and Sequential Models.}
Standard IRT and MF treat ability as static, whereas knowledge tracing (KT) models its temporal evolution.  
Bayesian Knowledge Tracing \citep{Corbett1994} and performance‑factor models are long‑standing baselines.  
Deep KT began with recurrent networks (DKT) \citep{Piech2015} and now includes memory networks \citep{Zhang2017} and attention mechanisms \citep{Ghosh2020}.  
However, simpler latent‑factor methods can rival deep KT when tuned well; \citet{Vie2019} outperformed DKT on Duolingo data.  
The NeurIPS 2020 Education Challenge attracted hybrid solutions mixing sequential models with meta‑learning and CF components \citep{Wang2021Competition}, underscoring a trend toward combining deep learning with probabilistic and factor models.

\paragraph{Positioning of Our Work}
We introduce the first discrete variational inference framework for predicting binary student responses, achieving state-of-the-art accuracy in both full and low-data regimes. Our approach builds on interpretable IRT and collaborative filtering, while remaining effective when student data is scarce. We evaluate on the largest open dataset of marked mathematics exams to date and show that variational inference improves predictions. We also show that class effects are important and show the model is able to uncover human interpretable skill dimensions.

\section{The Data}

Our data is provided by a large UK online learning platform that provides targeted mathematics revision for students. The data contains anonymous student scores (say 1 mark out of 3) on questions from GCSE (16 year old exams) mathematics mock papers they attempted. 
The mock exams are sat under formal exam conditions (closed book) by the students and marked accurately by qualified teachers, which is often peer-reviewed by other teachers.

\begin{figure}[t]
\centering
\includegraphics[width=0.5\textwidth]{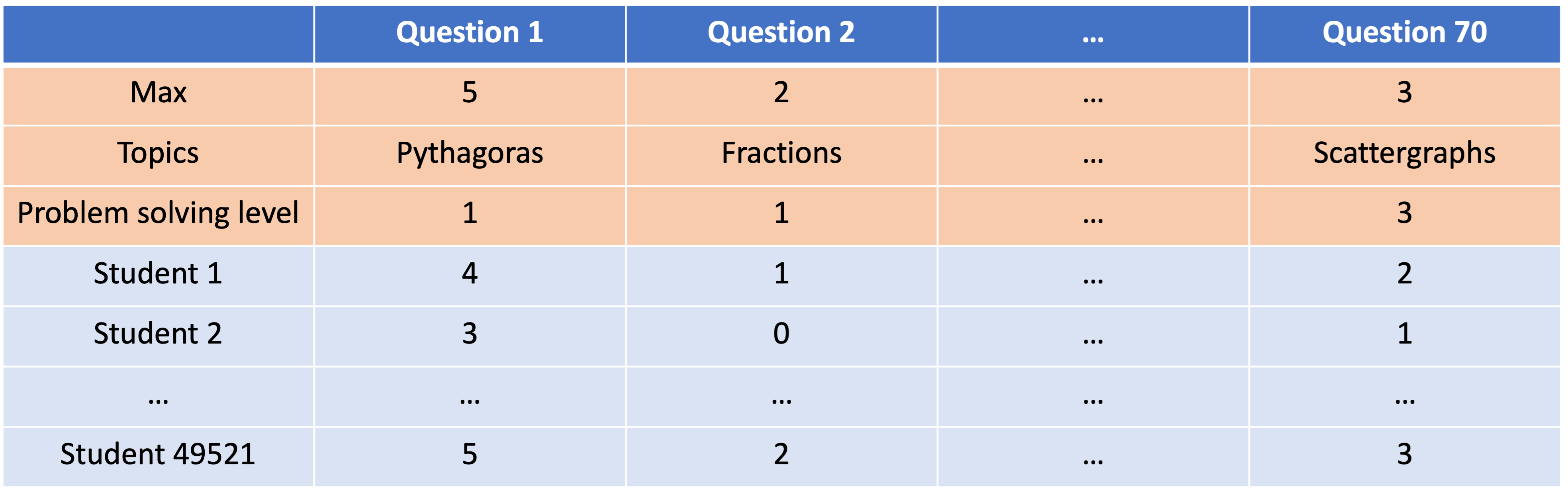}
\caption{Illustration of raw data}
\label{fig:raw_data_illustration}
\end{figure}

The raw dataset is illustrated in Figure \ref{fig:raw_data_illustration}, where the orange region represents the question metadata and the blue region represents students' raw scores.

\begin{figure}[t]
\centering
\includegraphics[width=0.5\textwidth]{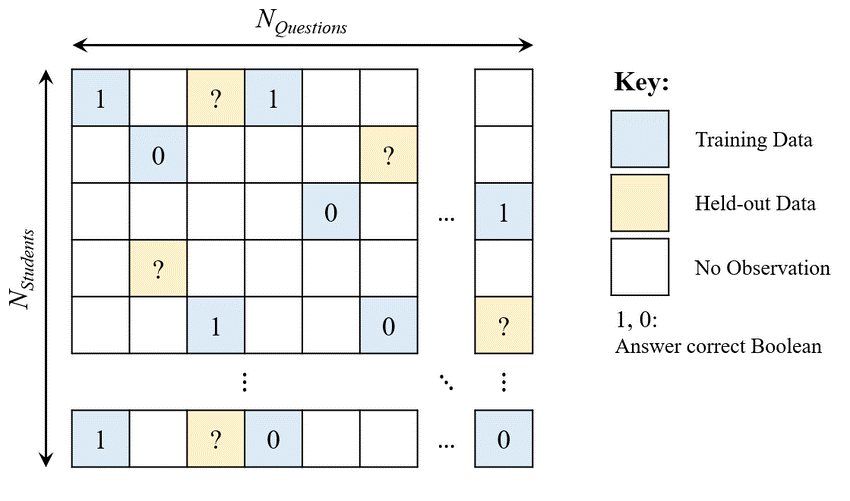}
\caption{Illustration of preprocessed data }
\label{fig:processed_data_illustration}
\end{figure}

Specifically, our models have been evaluated on data from the Edexcel GCSE 2017 Mathematics Paper 1, 2, and 3, containing 70 questions in total (24qns, 24qns, 22qns) and involving 49521 students. While most students have completed all 70 questions, some may have completed only one of the three papers, so there are fewer observed data points available for these students. The real data set may therefore be represented as a sparse matrix as illustrated in Figure \ref{fig:processed_data_illustration}, where the columns represent different questions and the rows represent different students, with unobserved data points spread over the data matrix. 
 
The paper will focus on probabilistic models on predicting correctness of students' responses and hence scores are binarised: a response is marked as 1 (correct) if the student achieved more than half the available marks for that question, and 0 (incorrect) otherwise.

\section{The Model}\label{sec:models}

We consider an Item Response Theory (IRT) model in the form of a hidden latent model, where student abilities and question difficulties can be regarded as the unobserved latent that are discovered from the parameters of the model. \citep{CTTIRT}. 

We look at separate formulations within the same family of models.

\subsection{Ability–Difficulty Model (IRT)}\label{subsec:irt}

The simplest model, the ability-difficulty models takes the form of a logistic function described by the following: 
\begin{equation} 
	p(y_{sq}=1 | \phi) = \frac{1}{1+e^{-(b_s + b_q)}} = \sigma(b_{s} + b_{q})
\label{eqt:logistic}
\end{equation}
This is also known as the Rasch model, where the probability that a question is answered correctly by a student is now calculated through two sets of latent parameters: $b_s$ representing the general ability of the student $s$ and $b_q$ representing the overall difficulty of the question $q$. A large positive $b_s$ implies a high-performing student whereas a large positive $b_q$ implies a simple question and vice versa. The parameters are optimised by stochastic gradient descent.

\subsection{The Interaction Model}\label{subsec:interactive}
On top of a general ability level, students may be strong in certain areas such as algebra and weak in others such as geometry. For example, two students who receive the same mark on an exam may have achieved it through different areas of mathematics.

We therefore extend the unidimensional IRT model, which can be extended to multidimensional (MIRT) models, to explore the interactions of multiple abilities, which is of particular interest in educational contexts in uncovering different skill dimensions possessed by students and required for solving specific questions \citep{MIRT2}.

We now have an overall ability level for each student, along with variables that can account for their strengths and weaknesses across different areas of mathematics. Similarly, we define an overall difficulty level for each question, as well as variables indicating which mathematical topics the question involves. These variables span \(n\) dimensions; for example, three dimensions could loosely correspond to algebra, statistics, and geometry.

Formally we have:
\[
\mathbf{b}_s = (b_{s0},\,1,\,b_{s1},\,\dots,\,b_{sD}), \quad
\mathbf{b}_q = (1,\,b_{q0},\,b_{q1},\,\dots,\,b_{qD})
\]
so that the probability becomes:
\begin{equation}
p(y_{sq}=1 \mid \phi)
=\sigma(b_{s0} + b_{q0} + \sum_{d=1}^D b_{sd}\,b_{qd})
\label{eqt:int_logistic}
\end{equation}

where $b_{s0}$ and $b_{q0}$ are synonymous with the overall student-ability and question-difficulty bias parameters in the Ability–Difficulty Model. \(b_{sd}\) represents the different proficiency a student may have in different areas, and \(b_{qd}\) is the question vector that contains the respective latents describing the question’s involvement in each of the cognitive areas \citep{cognitivediagnostic}.

\subsection{The Class Interaction Model}\label{subsec: Classinteractive}
The class interaction model is simply the interaction model with the individual skill levels of the students, $b_{sd}$, constrained to be the same for all students in a high school class. For example, all students in 11MA2 at say Springfield High School share the same $b_{sd}$. 

This model allows us to examine whether a teacher of a certain class may have taught a particular topic very well or poorly, or may not have covered another topic at all. 
 
The class interactive model is given by:
\[
\mathbf{b}_s = (b_{s0},\,1,\,b_{s1},\,\dots,\,b_{sD}), \quad
\mathbf{b}_q = (1,\,b_{q0},\,b_{q1},\,\dots,\,b_{qD})
\]
such that the probability becomes:
\begin{equation}
p(y_{sq}=1 \mid \phi)
= \sigma(b_{s0} + b_{q0} + \sum_{d=1}^D b_{cd}\,b_{qd})
\label{eqt:classint_logistic}
\end{equation}

Here, $b_{s0}$ and $b_{q0}$ are again synonymous with the overall student-ability and question-difficulty bias parameters from the Ability-Difficulty Model. However, each student in the same class $c$ now shares an identical $b_{cd}$, where $c \in {1, 2, \dots, n}$ represents the different areas of strength and weakness for class $c$, and there are $n$ unique classes.

\subsection{Variational Inference}\label{subsec: VI}
 
To be able to predict accurately for new platforms which have a low amount of students, we leverage variational inference treating student ability as a probability distribution. This approach allows us to capture uncertainty and individual variability in a principled way, rather than forcing a point estimate for each learner. What makes our application novel is that we derive a variational model within a discrete response setting, rather than the continuous domains found in the traditional variational inference literature. This requires special handling in both the formulation and optimisation.  

Specifically instead of point estimates we place independent Gaussian variational posteriors on every latent parameters:

\[
q(b_s) = \mathcal{N}(b_s \mid \mu_s, \sigma_s^2), \quad
q(b_{cd}) = \mathcal{N}(b_{cd} \mid \mu_{cd}, \sigma_{cd}^2),
\]
where the parameters \(\mu_s, \sigma_s^2, \mu_{cd}, \sigma_{cd}^2\) are optimised during training.

The variational objective is to maximise the evidence lower bound (ELBO), which balances the expected log-likelihood with a KL divergence regulariser:
\[
\text{ELBO} = \mathbb{E}_{q}[\log p(\mathbf{y} \mid \mathbf{b})] - \text{KL}(q(\mathbf{b}) \,\|\, p(\mathbf{b})).
\]

Algorithm 1 walks through our technique for the class interaction model, as this gains our greatest prediction accuracy. We are variational over student parameters only, including the class interaction terms, but all question parameters remain as point estimates.
The full novel derivation, sampling scheme, and closed-form expressions are provided in Appendix.

\begin{algorithm}[t]
\caption{Discrete variational inference for binary responses for the class interaction model}
\label{alg_dvi_simple}
\begin{algorithmic}[1]
\STATE Inputs, responses $y_{s,q}\in\{0,1\}$, answered item sets $\mathcal{Q}_s$, samples $M$, step size $\eta$, prior $p(b)=\mathcal{N}(0,1)$
\STATE Initialise, $\mu_s$, $\sigma_s$ for all students, $b_q$ for all question items
\STATE Initialise, $\mu_{cd}$, $\sigma_{cd}$ for all class interaction terms, $b_{qd}$ for all question interaction items

\REPEAT
  \STATE $\mathcal{L}\leftarrow 0$
  \FOR{each student $s$}
    \STATE draw $\epsilon_s^{(1)},\ldots,\epsilon_s^{(M)}\sim\mathcal{N}(0,1)$
    \STATE set $b_s^{(m)}\leftarrow \mu_s+\sigma_s\ \epsilon_s^{(m)}$
    \STATE draw $\epsilon_{cd}^{(1)},\ldots,\epsilon_{cd}^{(M)}\sim\mathcal{N}(0,1)$ for all $c,d$
\STATE set $b_{cd}^{(m)}\leftarrow \mu_{cd}+\sigma_{cd}\ \epsilon_{cd}^{(m)}$ for $m=1,\ldots,M$

\STATE $\widehat{\ell}_s\leftarrow \frac{1}{M}\sum_{m=1}^M\sum_{q\in\mathcal{Q}_s}\Big[y_{s,q}\log \sigma\!\big(b_s^{(m)}+b_q+\sum_{d} b_{c_s d}^{(m)}\, b_{q d}\big) + (1-y_{s,q})\log\!\big(1-\sigma\!\big(b_s^{(m)}+b_q+\sum_{d} b_{c_s d}^{(m)}\, b_{q d}\big)\big)\Big]$

\STATE $\text{KL}_s\leftarrow \log\!\frac{1}{\sigma_s}+\frac{\sigma_s^2+\mu_s^2}{2}-\frac{1}{2}$
\STATE $\mathcal{L}\leftarrow \mathcal{L}+\widehat{\ell}_s-\text{KL}_s$
  \ENDFOR
\STATE $\text{KL}_{\text{class}}\leftarrow \sum_{c,d}\Big(\log\!\frac{1}{\sigma_{cd}}+\frac{\sigma_{cd}^2+\mu_{cd}^2}{2}-\frac{1}{2}\Big)$
\STATE $\mathcal{L}\leftarrow \mathcal{L}-\text{KL}_{\text{class}}$
\STATE Update $\{\mu_s,\sigma_s\}$, $\{\mu_{cd},\sigma_{cd}\}$, $\{b_q\}$, and $\{b_{qd}\}$ by ascending $\nabla\mathcal{L}$ with gradient descent

\UNTIL{convergence}
\STATE \textbf{return} variational parameters $\{\mu_s,\sigma_s\}$, $\{\mu_{cd},\sigma_{cd}\}$ and item parameters $\{b_q\}$, and $\{b_{qd}\}$
\end{algorithmic}
\end{algorithm}


\section{Results and Discussion}\label{sec:eval}

The first thing to notice in our results shown in Table 1 is the \textbf{strong predictive power} of the single parameter ability difficulty model. 

\begin{table}[t]
    \centering
    \begin{tabular}{|l|c|c|}
        \hline
        \textbf{Results} & \textbf{1 Exam} & \textbf{3 Exams} \\ 
        \hline
        Students & 35{,}514 & 49{,}317 \\
        Questions & 24 & 70 \\
        Ability Difficulty & 80.8 & 80.6 \\
        Interaction (1D) & 80.7 & 80.7 \\
        Class Interaction & 82.1\,(18 Dims) & 81.9\,(35 Dims) \\
        InteractionVI (1D) & 80.9 & 80.7 \\
        Class InteractionVI & 82.15\,(18 Dims) & 80.8 \\
        \hline
    \end{tabular}
    \caption{Model performance across models and dataset sizes}
    \label{tab:exam-performance}
\end{table}

Our model achieved a precision of 80.8\% in our dataset, higher than the winning accuracy of 74.74\% in the NeurIPS 2020 Education Challenge~\cite{Wang2021Competition}. This indicates both the power of the model and the quality of the dataset.

To evaluate whether the ability–difficulty model could be outperformed, we swept over many dimensions for the interaction model. \textbf{Surprisingly}, we did not find any setting in which the interaction model exceeded the ability–difficulty model (see Appendix). We further evaluate the interaction model using synthetic data and demonstrate that, even when only weak topic‑level strengths and weaknesses are seeded into the simulated students, the model still succeeds in uncovering them to improve prediction accuracy. (see appendix)   

This leads us to our most significant insight for education:

\begin{center}
\textbf{Ability alone may be the sole determinant of success prediction.}
\end{center}

The concept that groups of students of the same overall ability may be strong in certain areas, such as algebra, and weak in others, such as geometry, may not be correct. For example, it is widely believed that students at a certain level (eg: Grade B) falls into certain groups. Some are strong in algebra and weak in geometry, and others have the opposite strengths. For example, studies such as Usiskin (1982) have shown that students often exhibit strong algebraic skills while remaining at relatively low levels of geometric reasoning \cite{usiskin1982van}. 

However, we are led to the conclusion that one parameter only, the ability of a student, is perhaps the only significant predictor of question success. 
 
This result could have far-reaching effects on education. If general mathematical ability is the only reliable predictor of exam performance, the curricula should shift the focus from isolated subtopics to building broad reasoning and problem-solving skills. Teacher training should emphasise fostering transferable cognitive skills that underpin mathematical ability across topics rather than improving individual topics in isolation. Intervention strategies should also target general mathematical development rather than narrow skill gaps in areas like algebra.  

While ability certainly dominates, there could well be other factors at play that are hard to find, although these must be much smaller. We worked hard to identify any such factors, exploring several avenues and model variations. We found one such factor that modestly improves prediction by 1.3 percent, only by introducing data on which high school class students were in.  The \textbf{class interaction model}, which groups students by their school classes, outperforms the ability-difficulty model. This suggests that the model can, for example, detect where a class teacher is particularly strong or weak in teaching certain topics, thereby enhancing its predictive power.

\subsection{Smaller Datasets Results: Discrete Variational Inference improves accuracy}

While we are fortunate to have access to a large dataset, many emerging educational platforms may not share this advantage. Predicting student ability in the early stages of such platforms presents a significant challenge due to limited data availability. It is within the class interaction model that the effectiveness of our discrete variational inference approach becomes most evident, as the presence of significant interaction effects allows our discrete variational model to achieve improved predictive accuracy, in low data settings.

To simulate this scenario, we evaluated our method on progressively smaller subsets of the dataset, shown in table 2. Notably, our novel discrete variational inference begins to yield significant improvements in predictive accuracy once we reach 15 percent of the full dataset, equivalent to approximately 5,000 students. Here the 0.7 percent increase is equivalent to 840 more correct predictions, a statistically significant increase at a 1 percent significance level (see appendix). 

\begin{table}[t]
    \centering
    \small
    \begin{tabular}{|l|c|c|}
        \hline
        \textbf{Dataset Size} & \textbf{CI (\%)} & \textbf{CIVI (\%)} \\ \hline
        Full Dataset (35,514) & 82.1 & 82.1 \\ \hline
        50\% Dataset (17,757) & 81.2 & 81.2 \\ \hline
        25\% Dataset (8,878) & 79.6 & 79.7 \\ \hline
        15\% Dataset (5,327) & 78.7 & 79.4 \\ \hline
    \end{tabular}
    \caption{Accuracy comparison using Standard Class Interaction (CI) and the Class Interaction Variational Inference model (CIVI) across dataset sizes for one exam}
\end{table}

And so here we see our new variational inference model can obtain greater prediction accuracy for smaller datasets. Note that these improvements are achieved by carefully fine-tuning the hyperparameters. Specifically, by initialising the standard deviations of student ability to values greater than 0.8, and by using a warm start for the parameters, initialised from the interaction model at its point of optimal predictive accuracy.

This suggests that, under appropriate conditions, our discrete variational inference can be a powerful tool for enhancing predictive performance in low-data regimes for educational platforms.
 
\section{Interpretability: Why does the class interaction model improve accuracy?}

We aim to investigate the factors contributing to the improvement in predictive accuracy observed when employing the Class Interaction (CI) model. To this end, we analyse the individual question embedding vectors $b_{qd}$ for all 24 questions generated by the best-performing model configuration (18 dimensions) on a single exam paper. Specifically, we compute the normalised cosine similarity between these vectors to explore the structure the model has learned.

Our analysis reveals that the model appears to identify latent thematic structures across the test items, effectively uncovering implicit \textit{question types}. For example, as illustrated in Figure 3, questions 4 and 20 exhibit high cosine similarity, and upon inspection, both are found to assess geometric reasoning. Similarly, questions 6 and 22 are both algebraic in nature, despite differing in difficulty. The model nonetheless assigns them closely aligned vectors, indicating it has captured an underlying algebraic theme. This pattern is repeated across several other question groupings and has been verified by fully qualified mathematics teachers. 

 The cosine similarity also appears to show the model has picked up on the fact that only strong ability classes are taught the hardest topics - elements of the $b_{qd}, d > 20$ vector are effectively encoding a latent query: 'Does this student belong to a high-ability class exposed to advanced topics?' and a corresponding element in the $b_{cd}$ answering yes or no.  
 
\begin{figure}[t]
\centering
\includegraphics[width=0.4\textwidth]{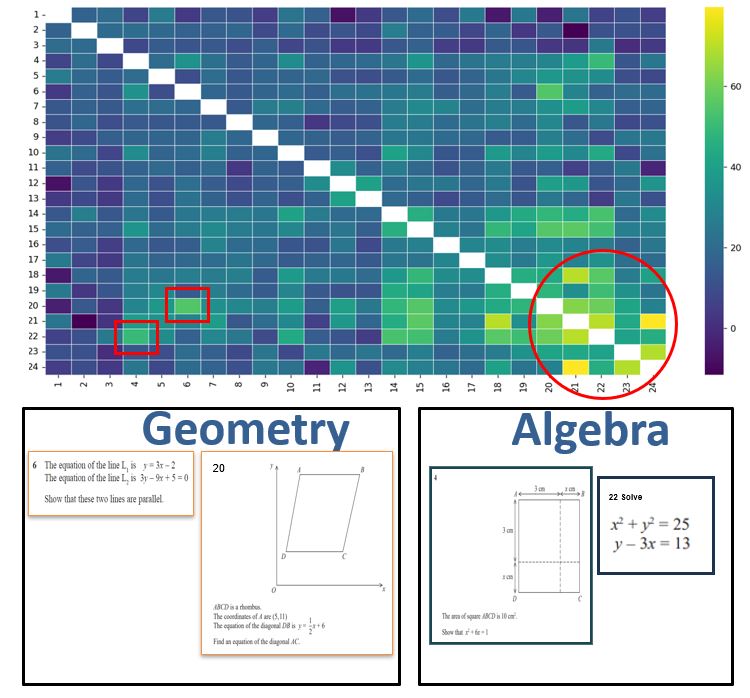}
\caption{Normalised cosine similarity of question vectors $b_{qd}$. The lower left  rectangle shows the model has discovered an underlying factor representing geometry, the other Algebra. The red circle shows the model has understood that only high ability classes actually get taught the hardest topics.}
\label{fig:TQ}
\end{figure}


\section{Active Learning} 

Having tackled prediction under data scarcity when a platform has few students, we now turn to the complementary problem: forecasting question success for entirely new students as soon as they join the system.

Thousands of new learners join educational platforms each day, yet most arrive with only a handful of answers on record. To cope with this data-sparse reality, we adopt pool-based active learning where the model estimates predictive uncertainty and queries the least-certain student–question pairs, updates its parameters and repeats. 

Our experiments partition the data into an initial labelled set, a large unlabelled pool of new students, and a held-out test set. We first fit the Ability–Difficulty model on a limited number of initial questions. We take its uncertainty over each unanswered question, and iteratively choose the next data point to select based on which questions predicted probability is closest to 0.5. Retraining after every batch of newly revealed labels prevents catastrophic forgetting and lets the model refine its ability estimates on the fly.

Results on 2000 unseen students in figure 4 show that active learning surpasses random querying after only ten questions, stabilising at a one-percentage-point accuracy lift and converging with roughly 30 questions per student—far fewer than passive baselines. Gains hold across low-, medium- and high-ability groups, with the largest relative improvement in the mid-ability cohort where prediction is hardest, demonstrating faster and more economical personalisation for intelligent tutoring systems.

\begin{figure}[t]
\centering
\includegraphics[width=0.5\textwidth]{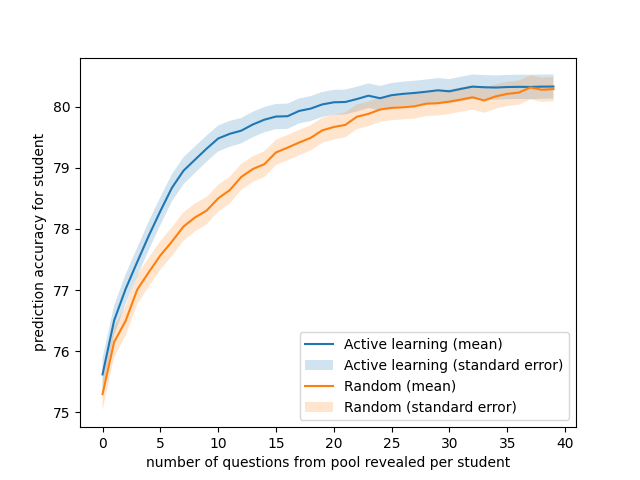}
\caption{Prediction accuracy of student averaged across all new students in data pool plotted against the number of questions per student so far revealed to the model}
\label{fig:AL_random_student_pred_acc}
\end{figure}


\section{Limitations and Future Directions}
\begin{itemize}
    \item \textbf{Static‐ability assumption.} The current models treat each student’s ability as constant over the exam window, overlooking learning or fatigue effects. In the future we will look to extend the discrete VI framework to temporal knowledge tracing so ability can evolve across sessions.
    
    \item \textbf{Single‐subject, UK‐centric dataset.} Our benchmark consists solely of GCSE mathematics sat by UK pupils, limiting generalisability to other subjects or regions. In the future we plan to replicate the study with multilingual, multi‐subject corpora to assess robustness and fairness across diverse cohorts.
    
    \item \textbf{Computational overhead of Variational Inference} Full Gaussian posteriors increase training time and memory relative to point‐estimate baselines. In the future we will explore lighter variational families and amortised inference to keep the method practical for large‐scale, real‐time tutoring platforms.
\end{itemize}

\section{Ethics Statement}
All data were de-identified before analysis, with no personal identifiers. We will release the dataset only in a form that preserves anonymity and complies with applicable data-protection obligations.

\section{Conclusion}\label{sec:conclusion}

We release the largest open, professionally marked GCSE mathematics dataset to date and show that a simple, interpretable ability–difficulty model already achieves strong accuracy. We introduce topic level skill variables in areas such as geometry and algebra for each student, but remarkably this fails to increase prediction power, suggesting that individual topic strengths contribute little once overall ability is accounted for. Many new platforms are unable to predict 
accurately due to lack of student and our main contribution is then a novel discrete variational-inference framework that delivers the biggest gains when data are scarce for new platforms and class information is available. 

Overall, we release a large open-source dataset, develop a series of models that uncover a potentially pivotal piece of evidence for education, and present a novel system for dealing with data scarcity in educational settings.

\bibliographystyle{plainnat}   
\bibliography{AAAI26}          

\appendix

\setlength\titlebox{1.2in} 

\title{Supplementary Material for Accurate Predictions in Education with Discrete Variational Inference}

\maketitle

\section{(A) Discrete Variational Inference Derivation}
 
We retain the Bernoulli likelihood in its exact discrete form, we do not introduce a continuous surrogate, we do not bound or augment the logistic term, and we optimise the expected Bernoulli log-likelihood under a Gaussian variational posterior. This keeps the derivation faithful to binary exam responses while remaining simple and scalable.

\paragraph{Prior and variational family}
We place independent standard normal priors on student abilities,
\[
 \text{  } b_s : \quad p(b_s) \sim \mathcal{N}(b_s ; 0, 1)
\]
For each observed pair \((s,q)\) with binary response \(y_{sq}\in\{0,1\}\),
\[
p\!\left( \{b_s\}_{s=1}^{S} \ =1,\middle|\, \{y_{s,q}\}_{s=1,q=1}^{S,Q} \right)
\approx \prod_{s=1}^{S} q_s(b_s),
\]
\[
\qquad
q_s(b_s) \,=\, \mathcal{N}\!\left(b_s \,;\, \mu_s, \sigma_s^{2}\right)
\]

\begin{align*}
\text{ELBO} &= \log p\left( \left\{ y_{s,q} \right\}_{s=1,q=1}^{S,Q} \middle| \left\{ b_q \right\}_{q=1}^Q \right) \\
&\quad - \text{KL} \left[ q\left( \left\{ b_s \right\}_{s=1}^S \right) \,\|\, p\left( \left\{ b_s \right\}_{s=1}^S \middle| \left\{ y_{s,q} \right\}_{s=1}^S \right) \right]
\end{align*}

\begin{align*}
\text{The evidence lower bound (ELBO) is } 
\end{align*}
\begin{align*}
&= \mathbb{E}_{q} \left[ \log p\left( \left\{ y_{s,q} \right\}_{s=1}^S \middle| \left\{ b_q \right\}_{q=1}^Q, \left\{ b_s \right\}_{s=1}^S \right) \right] \\
&\quad - \text{KL} \left( q\left( \left\{ b_s \right\}_{s=1}^S \right) \,\|\, p\left( \left\{ b_s \right\}_{s=1}^S \middle| \left\{ y_{s,q} \right\}_{s=1}^S \right) \right)
\end{align*}

\begin{align*}
= \sum_{s=1}^S \mathbb{E}_{q_s(b_s)} \bigg[ &\log p\left( \left\{ y_{s,q} \right\}_{q=1}^Q \middle| \left\{ b_q \right\}_{q=1}^Q, \left\{ b_q \right\}_{s=1}^S \right) \\
&\quad - \text{KL} \left( q_s(b_s) \,\|\, p_s(b_s) \right) \bigg]
\end{align*}

\begin{align*}
= \mathbb{E}_{q_s(b_s)} \bigg[ 
\sum_{q=1}^{Q_s} \log \bigg( &
\left( \frac{e^{b_s + b_q}}{1 + e^{b_s + b_q}} \right)^{y_{sq}} 
\left( \frac{1}{1 + e^{b_s + b_q}} \right)^{1 - y_{sq}} 
\bigg)
\bigg] \\
&\quad - \text{KL}\left( \mathcal{N}(b_s ; \mu_s, \sigma_s^2) \,\|\, \mathcal{N}(b_s ; 0, 1) \right)
\end{align*}
We keep the Bernoulli likelihood exact and move stochasticity to the continuous latent by
\[
\quad\quad
b_s^{(m)} = \mu_s + \sigma_s \, \epsilon_m, \quad \epsilon_m \sim \mathcal{N}(0, 1)
\]
A Monte Carlo estimator with \(M\) samples gives
\begin{align*}
\text{ELBO} \approx 
\frac{1}{M} \sum_{s=1}^S \sum_{m=1}^M \sum_{q=1}^{Q_s}
\bigg[
  y_{sq} \left( b_s^{(m)} + b_q \right)
  - \log\left(1 + e^{b_s^{(m)} + b_q}\right)
\bigg]
\end{align*}
\begin{align*}
-  \sum_{s=1}^S \text{KL} \left( q(b_s) \,\|\, p(b_s) \right)
\end{align*}

\[
\text{where} \quad b_s^{(m)} = \mu_s + \sigma_s \epsilon_s^{(m)}, 
\quad \epsilon_s^{(m)} \sim \mathcal{N}(0, 1)
\]

\[
\text{and} \quad \text{KL} \left( \mathcal{N}(x_j ; \mu_1, \sigma_1^2) \,\|\, \mathcal{N}(x_j ; \mu_2, \sigma_2^2) \right)
= \log \frac{\sigma_2}{\sigma_1} 
+ \frac{\sigma_1^2 + (\mu_1 - \mu_2)^2}{2\sigma_2^2} - \frac{1}{2}
\]

We optimise \(\{\mu_s,\sigma_s\}\) and \(\{b_q\}\) by gradient ascent .  
In practice we initialise \(\sigma_s\) above a small positive floor to reflect uncertainty under scarce data.
\begin{flushright}
$\blacksquare$
\end{flushright}

\subsection{B Two–proportion $z$-test for the significance of the class interaction model with variational inference improvement over standard class interaction model for 15 percent of the dataset ($\alpha = 0.01$)}
\begin{align*}
n_1 &= n_2 = 120\,000, (number of questions) &
x_1 &= 94\,440, & \\
x_2 &= 95\,280, \\[2pt]
\hat p &= \frac{x_1 + x_2}{n_1 + n_2} = 0.7905, \\[2pt]
\mathrm{SE} &= \sqrt{\hat p(1-\hat p)\!\left(\tfrac1{n_1} + \tfrac1{n_2}\right)}
           = 0.00166, \\[2pt]
z &= \frac{p_2-p_1}{\mathrm{SE}}
   = \frac{0.794 - 0.787}{0.00166}
   = 4.21, \\[2pt]
z_{0.99} &= 2.576, \qquad
p \approx 2.5\times10^{-5}\;{<}\;0.01.
\end{align*}
\textbf{Result:} \(|z| > z_{0.99}\); \\the 0.7 pp improvement is statistically significant at the 1 \% level.
\section{(C) Simulation Demonstrating Interaction Model Recovery}

To demonstrate that our interaction model can successfully learn underlying skill dimensions when they exist, we conducted a controlled simulation with synthetic data using the same scale as our real dataset \\(40,000 students and 24 questions).

We generated student and question latent variables using the following formulation:
\begin{itemize}
    \item Student ability biases $b_s \sim \mathcal{N}(0, 1)$
    \item Question difficulty biases $b_q \sim \mathcal{N}(-3, 1)$
    \item Student skill vectors $x_s \sim \mathcal{N}(0, 1)$ in $D=1$ dimension
    \item Question skill demands $x_q \sim \mathcal{N}(0, 1)$ in $D=1$ dimension
\end{itemize}

We simulated binary responses using this model and trained both our ability–difficulty model and the 1D interaction model on the resulting dataset. \\
As expected, the interaction model achieved higher prediction accuracy:
\begin{itemize}
    \item Ability–difficulty model: \textbf{95.54\%}
    \item 1D interaction model: \textbf{96.20\%}
\end{itemize}

This result confirms that our interaction model can recover underlying topic-level structure and yield improved predictions when meaningful interactions between student strengths and question demands are present.

\end{document}